\documentclass[10pt,twocolumn,letterpaper]{article}

\usepackage{cvpr}              %
\usepackage{subfiles}

\makeatletter
\renewcommand\paragraph{\@startsection{paragraph}{4}{\z@}{-0.8mm}{-2.8mm}{\bf\itshape\@adddotafter}}
\makeatother
\usepackage{flushend}
\usepackage{xcolor}
\usepackage{mdframed}

\newcommand{\comment}[1]{}
\newcommand{\mparagraph}[1]{\vspace{1mm}\noindent{\textbf{#1.}\hspace{1mm}}}

\usepackage{import} %
\graphicspath{{figs/}}

\usepackage[ruled,noend]{algorithm2e} %

\SetAlFnt{\small}
\SetAlCapFnt{\small}
\SetAlCapNameFnt{\small}
\SetAlCapHSkip{0pt}
\IncMargin{-\parindent}
\usepackage{ifpdf}
\usepackage{graphicx}
\ifpdf
  
\else
\fi
\usepackage{color}
\definecolor{cyan}{cmyk}{1,0,0,0}
\definecolor{darkgreen}{rgb}{0,0.5,0}
\definecolor{orange}{rgb}{1,0.5,0}
\definecolor{magenta}{cmyk}{0,1,0,0}
\definecolor{darkyellow}{cmyk}{0,0,0.75,0}
\definecolor{gray}{rgb}{0.8,0.8,0.8}

\usepackage{amsmath} %

\usepackage[toc,page]{appendix} %
\usepackage{wrapfig} %
\usepackage{comment}
\usepackage{bm}
\usepackage{cuted} %
\usepackage{lipsum} %
\usepackage{mathtools} %
\usepackage{algorithmicx}
\usepackage{algpseudocode}
\usepackage{nicefrac}

\usepackage{gensymb}
\usepackage{float}
\usepackage{soul}
\usepackage{array}
\usepackage{multirow}
\usepackage{hhline}
\usepackage{setspace}
\usepackage{nicefrac}

\algdef{SE}[DOWHILE]{Do}{doWhile}{\algorithmicdo}[1]{\algorithmicwhile\ #1}%

\makeatletter
\renewcommand{\ALG@beginalgorithmic}{\small}
\makeatother

\newcommand{\DELETE}[1]{} %
\newcommand{\IGNORE}[1]{}
\usepackage{datenumber}
\usepackage{calc}
\usepackage[mmddyyyy]{datetime}

\newcounter{datetoday}
\newcounter{diffyears}
\newcounter{diffmonths}
\newcounter{diffdays}

\newcommand{\difftoday}[3]{%
      \setmydatenumber{datetoday}{\the\year}{\the\month}{\the\day}%
      \setmydatenumber{diffdays}{#1}{#2}{#3}%
      \addtocounter{diffdays}{-\thedatetoday}%
      \ifnum\value{diffdays}>0
        \def\diffbefore{}%
        \def\diffafter{left}%
      \else
        \def\diffbefore{}%
        \def\diffafter{ago}%
        \setcounter{diffdays}{-\value{diffdays}}%
      \fi
      \setcounter{diffyears}{\value{diffdays}/365}%
      \setcounter{diffdays}{\value{diffdays}-365*\value{diffyears}}%
      \setcounter{diffmonths}{\value{diffdays}/30}%
      \setcounter{diffdays}{\value{diffdays}-30*\value{diffmonths}}%
      \diffbefore
      \ifnum\value{diffyears}=0
      \else
        \ifnum\value{diffyears}>1
            \thediffyears\space years,
        \else
            \thediffyears\space year,
        \fi
      \fi
      \ifnum\value{diffmonths}=0
      \else
        \ifnum\value{diffmonths}>1
            \thediffmonths\space months
        \else
            \thediffmonths\space month
        \fi
      \fi
      \ifnum\value{diffdays}=0
      \else
        \ifnum\value{diffdays}>1
            \thediffdays\space days
        \else
            \thediffdays\space day
        \fi
      \fi
      \diffafter
}

\definecolor{cvprblue}{rgb}{0.21,0.49,0.74}
\usepackage[pagebackref,breaklinks,colorlinks,citecolor=cvprblue]{hyperref}
\usepackage[accsupp]{axessibility}
\def\paperID{3029} %
\def\confName{CVPR}
\def\confYear{2024}

\title{OmniSDF: Scene Reconstruction using\\Omnidirectional Signed Distance Functions and Adaptive Binoctrees}

\author{
\begin{tabular}{c c c c c}
Hakyeong Kim & Andreas Meuleman & Hyeonjoong Jang & James Tompkin & Min H. Kim \\
KAIST & INRIA & KAIST & Brown University & KAIST \\
\end{tabular}
}

\begin{document}
\maketitle
\begin{abstract}
\noindent
We present a method to reconstruct indoor and outdoor static scene geometry and appearance from an omnidirectional video moving in a small circular sweep. This setting is challenging because of the small baseline and large depth ranges, making it difficult to find ray crossings. To better constrain the optimization, we estimate geometry as a signed distance field within a spherical binoctree data structure and use a complementary efficient tree traversal strategy based on a breadth-first search for sampling. Unlike regular grids or trees, the shape of this structure well-matches the camera setting, creating a better memory-quality trade-off. From an initial depth estimate, the binoctree is adaptively subdivided throughout the optimization; previous methods use a fixed depth that leaves the scene undersampled. In comparison with three neural optimization methods and two non-neural methods, ours shows decreased geometry error on average, especially in a detailed scene, while significantly reducing the required number of voxels to represent such details.
\end{abstract}

\vspace{-0.1cm}
\section{Introduction}
\label{sec:intro}
\vspace{-0.15cm}
When reconstructing the geometry of a static unbound scene, say an outdoor space, most image-based multi-view reconstruction pipelines use perspective cameras. 
Given their limited field of view, they suffer from high data acquisition costs as many images must be captured and calibrated.
But, given their physically-compact size, it is possible to move to different spatial sampling positions and capture many light ray crossings from which to accurately constrain the surface geometry location.
To solve for the location, modern auto-differentiation systems let us flexibly implement algorithms like sphere tracing with neural signed-distance functions (SDFs).

Given enough ray crossings, these algorithms show promise in capturing geometry from real-world perspective images, both in geometry detail~\cite{li2023neuralangelo} and in large-scale structures~\cite{sun2022neural}, thanks to the robust fitting properties of neural networks.
If memory is not a concern, we can exploit voxel grid structures for rapid position indexing~\cite{sun2022neural, li2023neuralangelo}, but unbound scenes require careful memory consideration.
\begin{figure}[pt]
	\centering
	\includegraphics[width=\linewidth]{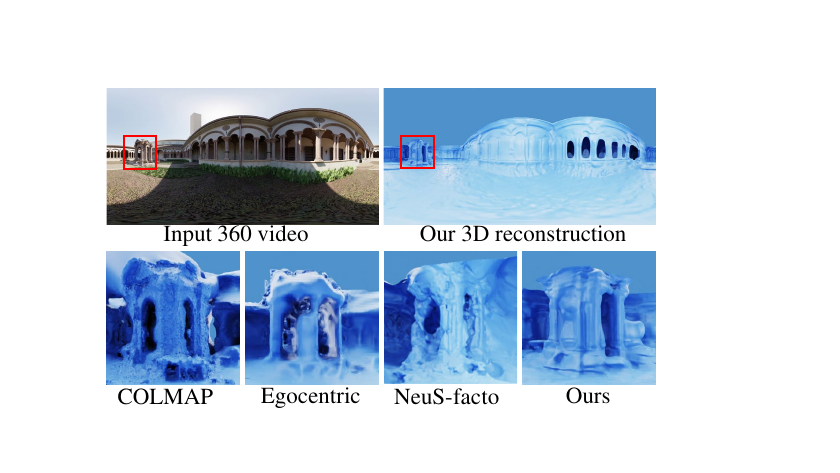}%
	\vspace{-3mm}%
	\caption[Sampling strategies]{\label{fig:abstract}%
We introduce a memory-efficient neural 3D reconstruction method tailored to work with short egocentric omnidirectional video inputs. The geometry is estimated using a signed distance field and a novel adaptive spherical binoctree data structure subdivided through iterative optimization. We show that our method outperforms other state-of-the-art 3D reconstruction methods in balancing detail and memory cost~\cite{schonberger2016structure, jang2022egocentric, Yu2022SDFStudio}. 
}
\vspace{-5mm}
\end{figure}

To help overcome the acquisition cost, one alternative is to use an omnidirectional camera that captures a view of all angles of the surroundings. This is a critical capability when comprehensive spatial understanding is necessary, such as in virtual reality and robotic navigation.
But, now a wider field of view is represented within a limited camera sensor resolution, such that the angular extent of each pixel is typically greater than in a perspective camera ($\approx 5\times$).
To gain parallax for distance estimation, suppose we additionally consider a limited spatial sampling setting, e.g., an omnidirectional video of a circular camera motion---this would be convenient for data capture~\cite{guo2023streetsurf}.
But, as the baseline between spatial positions is small with respect to the large range of scene distances, and as each pixel covers a greater angular extent, it is difficult to accurately constrain the location of the geometry or to reproduce its fine detail.
These factors decrease the effectiveness of ray marching techniques and, within an optimization, often necessitate an impractical number of samples for successful training convergence.
This issue is further compounded by rapid and large changes in depth in outdoor scenes---a difficult function to fit---that typically leads to a loss of detail in near-field geometry.
In terms of memory, naive grids do not scale well to unbound scenes, and approaches tailored to small-baseline omnidirectional cameras should be more efficient.

To address these challenges, we propose a neural 3D reconstruction method specifically designed for short egocentric omnidirectional video inputs. 
Central to our approach is an \emph{adaptive} spherical binoctree formed of spherical voxels (`sphoxels'). 
This dynamically subdivides from coarse to fine granularity through an iterative optimization, such that both ray samples and memory can focus on areas with more detail. 
Within the spatially-varying subdivisions of the leaf nodes, each sphoxel is fine-tuned to optimize spatial resolution and prioritize sampling in areas closer to the observed surfaces.
Beyond this, we also provide two practical solutions: an efficient tree traversal strategy that manages the inconsistent number of child nodes based on breadth-first search, and an intersection test method to accommodate uneven shapes of the sphoxels in the binoctree structure.

Binoctrees were originally designed to subdivide a spherical space while compensating for voxel elongation that occurs at far distances~\cite{jang2022egocentric}. 
However, this previous approach does not adapt the structure: not by the achievable depth accuracy, which can be estimated given camera poses, nor by any potential initial depth value, nor by the depth during optimization itself. 
We show that our approach improves upon an existing binoctree-based reconstruction method, reducing error by {38\%} on average. We also show improvement over a recent neural-field-based SDF reconstruction method that does not efficiently partition space within large-scale scenes~\cite{wang2021neus}.
Overall, our approach allows for more efficient and accurate 3D surface reconstruction of large-scale unbounded scenes using omnidirectional video inputs. Our method more effectively reconstructs the 3D environment from egocentric video inputs, addressing challenges in neural geometry with omnidirectional cameras.
Our code is freely available for research purposes\footnote{\url{https://vclab.kaist.ac.kr/cvpr2024p2/}}.

\vspace{-0.1cm}
\section{Related Work}
\label{sec:relatedwork}
\vspace{-0.15cm}

\mparagraph{360$\degree$ photography}
Scene understanding from 360$\degree$ images has been actively explored together with the growth of the VR/AR industry.
Using VR as a motivating application, several works focus on image-based rendering of 360$\degree$ images from novel viewpoints~\cite{attal2020matryodshka, bertel2020omniphotos, choi2023balanced}. These works adopt convolutional neural networks to construct multi-sphere images, proxy geometry from learned optical flow, and neural radiance fields. However, these works do not explicitly aim to understanding scene geometry and, therefore, suffer distortion when rendering trajectory deviates from training trajectories.
Other work estimates dense depth and normal maps for corresponding 360$\degree$ RGB images for inverse rendering~\cite{li2022phyir}. 
However, this is limited to indoor scenes, and output depth maps require post-processing steps for explicit mesh extraction from obtained point clouds~\cite{kazhdan2006poisson, kazhdan2013screened}.
On the other hand, our work focuses more on the geometrical understanding of spherical scenes by directly estimating the accurate signed distance function (SDF).

Jang et al.~\cite{jang2022egocentric} share our goal in reconstructing a mesh of large-scale outdoor scenes from omnidirectional video. 
They use a binoctree structure to efficiently subdivide the unbounded space, then fuse truncated SDF (TSDF) values directly. This makes the output mesh resolution depend upon the voxel grid resolution, which is fixed from an initial depth estimate (e.g., from a pre-trained depth estimation network). 
Any errors in the initial depth also cause errors in fusion.
Our work adopt a neural implicit function for surface representation, using an adaptive binoctree to guide sampling during reconstruction.
This lets us refine any incorrect initial depth estimates, and does not require parameter tuning for fusion weights and truncation thresholds.

\mparagraph{Scene-scale surface reconstruction}
Structure-from-motion (SfM) and multi-view stereo (MVS) techniques have successfully reconstructed city-scale scenes from multi-view images~\cite{agarwal2011building, schonberger2016structure, knapitsch2017tanks}. 
However, these methods represent the 3D geometry with sparse point clouds that require extra steps to extract a surface mesh~\cite{kazhdan2006poisson,kazhdan2013screened}.
With the emergence of radiance fields as a 3D scene representation~\cite{gao2022nerf}, follow-up multi-view reconstruction studies have tackled outdoor scenes from object-centric unbounded scenes~\cite{barron2022mip, zhang2020nerf++} to larger scenes approaching urban scale~\cite{tancik2022block, rematas2022urban, gao2022nerf, meuleman2023localrf, muller2022instant, wu2023scanerf}.
Radiance-field-based methods work well for novel view synthesis but often fail to estimate correct geometries from sparse views. 
Tang et al.~\cite{tang2023delicate} refine an extracted surface mesh to acquire delicate geometry and a texture map from radiance fields. However, they do not aim to reconstruct scene-scale geometry.

Neural implicit representations using SDFs show potential for image-based 3D reconstruction through several works~\cite{wang2021neus, NEURIPS2020_1a77befc, oechsle2021unisurf, yariv2021volume, li2022vox}. These methods optimize an implicit function within the volumetric rendering pipeline by designing a density function derived from the estimated SDF.
Recent works~\cite{sun2022neural, li2023neuralangelo, guo2023streetsurf} extend the reconstruction scale to unbounded outdoor scenes.
To handle optimization difficulties and preserve high-fidelity details, 
Sun et al.~\cite{sun2022neural} use a sparse voxel grid built with SfM depth for the sampling, 
Li et al.~\cite{li2023neuralangelo} adopt numerical gradients and level-wise optimization of hash grid features~\cite{muller2022instant}, and Guo et al.~\cite{guo2023streetsurf} used cuboid-shaped hash grids and long-trajectory initialization schemes to handle unbounded street scenes.

Our work is built upon NeuS~\cite{wang2021neus} to take advantage of a neural implicit representation; this embeds the geometry reconstruction optimization procedure inside a differentiable image-based rendering pipeline.
However, our work differs from previous research in how we interpret the reconstruction space. To handle narrow-baseline outward-facing scenes, we use an adaptive binoctree within spherical space for sampling. 
This is different from common approaches that use Cartesian coordinates for space sampling.

\mparagraph{Grid-based training strategies}
Voxel grids are widely used to accelerate rendering at the cost of memory. 
NSVF~\cite{liu2020neural} uses a sparse voxel grid that adaptively prunes itself to narrow the sampling range. DVGO~\cite{sun2022direct} and Plenoctrees~\cite{yu2021plenoctrees} store density and appearance encodings to speed up training time and rendering.
Instant-NGP~\cite{muller2022instant} uses a hash table across multiple grid levels.
These methods build a grid in NDC space based on Cartesian coordinates. 
Cartesian voxel grids assume that regions of interest are spread evenly within a cube and that sampling of the space with cameras follows similarly. 
However, for narrow-baseline outward-facing sampling (sometimes called egocentric), we need a novel design that can reduce sampling to plausible regions while encompassing unbounded scene points.
EgoNeRF~\cite{choi2023balanced} uses a spherical feature grid as a neural scene representation to optimize a distant environment but focuses on novel view synthesis and does not reconstruct a surface.
Our adaptive spherical binoctree method for omnidirectional inputs recovers 3D geometry with improved details and convergence speed over Cartesian grids.

\vspace{-0.1cm}
\section{Omnidirectional SDF Reconstruction}
\label{sec:ourmethod}
\vspace{-0.15cm}

Given an omnidirectional video captured in a circular sweep, our goal is to optimize a neural SDF within a spherical space from which to reconstruct a 3D surface mesh.
Our reconstruction algorithm starts from training NeuS~\cite{wang2021neus} along with an initial binoctree constructed by the per-frame initial depth from a pre-trained spherical depth estimation network~\cite{jang2022egocentric}.

Then, we use step-wise sampling strategies with adaptive binoctree subdivision to optimize both the spherical grid and the neural networks with an image-based reconstruction loss.
The core ideas of our approach are: 1) voxel-guided sampling with a sphere-shaped grid that subdivides the reconstruction space, which is specialized for memory efficiency, and 2) online and iterative refinement of grid structures based on the intermediate results.

\subsection{Preliminaries}

\mparagraph{Implicit surface rendering}
We define a surface~$\mathcal{S}$ as the zero-level set of a signed distance function $f:{\mathbb{R}^3} \to \mathbb{R}$:
\begin{align}\label{eq:surface}
	\mathcal{S} = \left\{ {\mathbf{x} \in {\mathbb{R}^3}|f\left( \mathbf{x} \right) = 0} \right\}.
\end{align}
We use two neural networks to estimate $\mathcal{S}$~\cite{NEURIPS2020_1a77befc,wang2021neus}: 1) Given a 3D point $\mathbf{x}$, $F_\theta^{\text{SDF}} : (\mathbf{x}) \rightarrow (d,\zeta)$ estimates the signed distance $d$ to the surface at scene point $\mathbf{x}$, and additionally outputs a 256-dimensional feature vector $\zeta$. 2) $F_\theta^{\text{c}} : (\mathbf{x},\mathbf{n},\mathbf{v},\zeta) \rightarrow (c)$ estimates the view-dependent appearance $c$ at scene point $\mathbf{x}$ using the normal $\mathbf{n} \left( t \right)$ to the SDF as its derivative. To render an image, we sample points along a ray $\left\{ {{\mathbf{x}}\left( t \right) = {\mathbf{o}} + t{\mathbf{v}}|t \geqslant 0} \right\} $, where $\mathbf{o}$ is the ray origin and $\mathbf{v}$ is the ray direction. Then, we compute Wang et al.'s~\cite{wang2021neus} unbiased and occlusion-aware weights $w(t)$ from the estimated SDFs to accumulate the final output color $C$ by volume rendering:
\begin{equation}\label{eq:color}
	C\left( {{\mathbf{o}},{\mathbf{v}}} \right) = \int_{\textup{near}}^{\textup{far}} {w\left( t \right)c\left( {{\mathbf{x}}\left( t \right),{\mathbf{n}}\left( t \right),{\mathbf{v}},\zeta \left( t \right)} \right)dt}.
\end{equation}

\mparagraph{Binoctree} 
The spherical space is defined by a spherical binoctree~\cite{jang2022egocentric}: a sphere-shaped octree structure that subdivides a sphere in radial ($r$), polar ($\theta$) and azimuthal ($\phi$) directions. We call each cell a sphoxel. This structure allows binary subdivisions in the radial direction to prevent sphoxel elongation as $r$ increases, and is bound by near and far spheres.
Each sphoxel stores its boundary in spherical coordinates
[$r_\text{min}, r_\text{max}, \theta_\text{min}, \theta_\text{max}, \phi_\text{min}, \phi_\text{max}$]
and its index. We store sphoxel indices for each sampling stage and traverse the binoctree to efficiently fetch intersecting sphoxels.

\mparagraph{Preprocessing} 
Following Jang et al.~\cite{jang2022egocentric}, this stage estimates camera poses and initializes the spherical binoctree from a depth estimate.
We estimate depth for the initial tree construction using Jang et al.'s public trained model and code. We use structure from motion to solve for per-frame camera poses from the omnidirectional video (OpenVSLAM~\cite{sumikura2019openvslam}). Given the poses, we extract a dense 3D point cloud for each frame using spherical rectification~\cite{li2008binocular} and a fine-tuned disparity estimation network~\cite{teed2020raft}.

Using per-frame camera poses and calibrated depth maps as input, we form a global coordinate frame in metric units for the 3D point cloud. We place the origin at the center of the cameras and set the near sphere bound to contain all the camera positions.
To represent the color of scene elements at infinity, such as the sky, we place a final sphere at the far plane with view-independent color decoded by a background MLP. We set the far sphere bound per scene to fit the recovered point cloud inside the sphere. Before binoctree construction and the neural network training, we scale both binoctree and point cloud coordinates into a unit sphere, and store the coefficient for later metric re-scaling.

For the initial binoctree structure, we use the scaled initial 3D point cloud to subdivide the tree to its finest level until all sphoxels reach a minimum solid angle $\alpha_{\text{min}}$. 
We derive this size per scene from the baseline and image sizes, and set $\alpha_{\text{min}}$ between $5e^{-4}$ and $1e^{-3}$.

\subsection{Sphoxel Intersection and Surface Existence}
\label{sec:intersection}
\begin{figure}[pt]
	\centering
	\includegraphics[width=1\linewidth]{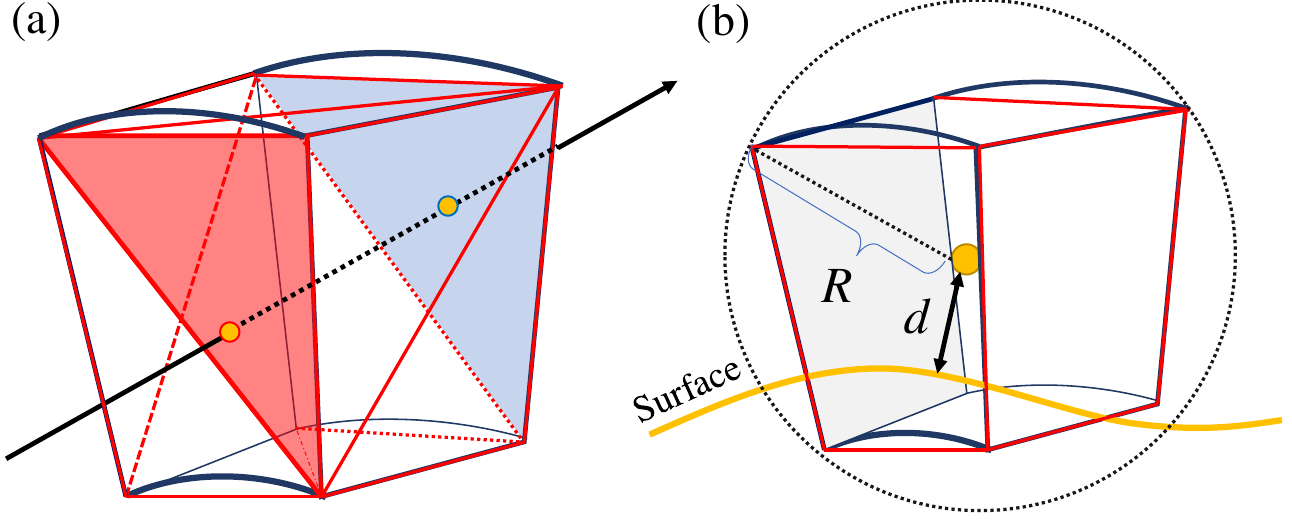}%
	\vspace{-3mm}
	\caption[Sphoxel intersection]{
		(a) Cuboid approximation of a sphoxel for intersection testing. Ray-triangle intersection tests occur for all triangles that comprise cuboid faces.
		(b) Illustration of rubrics for surface detection in sphoxels. A surface exists in a sphoxel if  $\|d\| < R$. 
		\label{fig:sphoxel_surface_intersection}}
	\vspace{-3mm}
\end{figure}

Unlike the previous binoctree method that directly integrates the SDF from depth maps~\cite{jang2022egocentric}, we optimize a hybrid binoctree/MLP SDF representation.
We use the spherical binoctree only to guide sampling within the optimization, providing an efficient subdivision for large spherical spaces. Given the irregular shape of sphoxels, we cannot adopt conventional Cartesian grid-based algorithms for ray intersection and occupancy search~\cite{liu2020neural, yu2021plenoctrees}. Thus, we suggest a novel intersection and surface existence test for sphoxels.

Each sphoxel is formed of two curved surfaces that each lie on a sphere defined by a near and far distance along radius $r$, and four surfaces by the intersection of those spheres with bounding planes formed at $\theta_\text{min},\theta_\text{max},\phi_\text{min},\phi_\text{max}$ (Figure~\ref{fig:sphoxel_surface_intersection}).
Only two of these surfaces are simple in that they can be approximated by two triangles.
Therefore, we cannot adopt common AABB cube intersection algorithms for sphoxel intersections.
We re-define intersection for sphoxels by approximating them as trapezoidal prisms (square-based frusta) that share the same vertices (Figure~\ref{fig:sphoxel_surface_intersection}(a)).
Using the triangle-ray intersection algorithm, a ray intersects a sphoxel if we can detect intersections between a ray and a pair of triangles---one front-facing, one back-facing---that form the faces.
We implement a CUDA kernel that fetches all intersecting sphoxels denoted by $V$.
Since the number of marked voxels easily exceeds 10\,k, we traverse the binoctree in a breadth-first order to relieve the bottleneck caused by unnecessary intersection computations.

When an angular subdivision spans a pole, then the binoctree around the pole forms a spherical sector (a cone and spherical cap) that contains the polar axis.
We divide the spherical sector around the polar axis and approximate these sphoxel segments as triangle-based frusta.

Computing sphoxel intersections lets us detect where the current surface is, but we need another way to mark sphoxels with likely surface-crossings to be sampled in future iterations.
For this, we define sphoxel center and radii. Since we compute intersections from the square-frusta approximation, we also define the center and radii from approximate geometry.
A sphoxel's center is the mean position of its constituent eight vertices, and its radius is the distance from its center to the farthest vertex. 
Figure~\ref{fig:sphoxel_surface_intersection}(b) compares the inferred absolute SDF value $\|d\|$ at the sphoxel center with the corresponding sphoxel radius $R$. 
We assume the surface passes through the sphoxel if $\|d\|$ is smaller than $R$.

\subsection{Spatial Sampling Types}
\label{sec:sampling}
\begin{figure}[pt]
	\centering
	\vspace{-5mm}
	\includegraphics[width=1\linewidth]{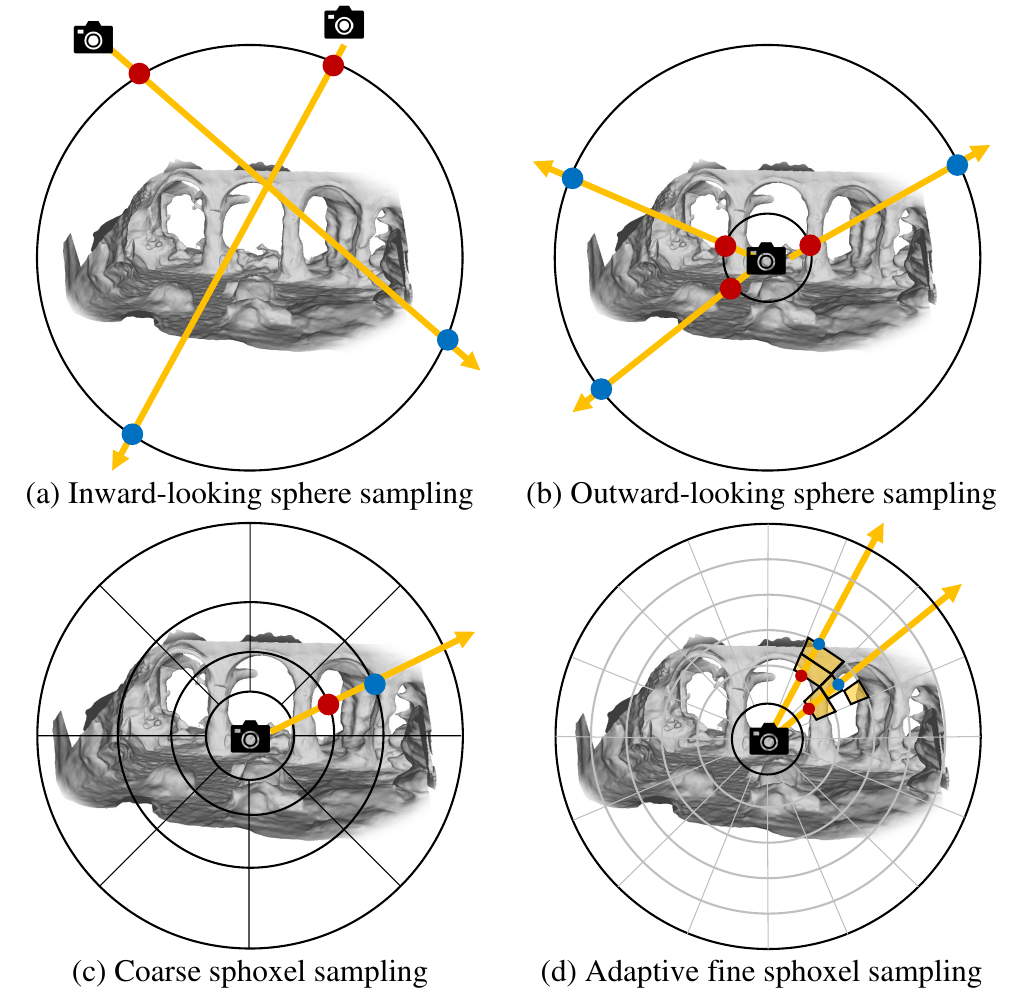}%
	\vspace{-3mm}
	\caption[Sampling types]{Sampling strategies on a unit sphere. 
	(a) Sphere sampling range for the exocentric inward-looking model.
	(b)--(d) Sampling for egocentric data using a coarse-to-fine binoctree.
		\label{fig:sampling_types}}
	\vspace{-4mm}
\end{figure}

We use three different levels of space sampling.

\mparagraph{Whole-space spherical sampling}
This occurs along each input camera ray between its intersections with the near and far bounding spheres.
Since we assume that all cameras are located inside the near-bound sphere, rays uniquely intersect with both spheres.
Figure~\ref{fig:sampling_types}(a) illustrates how this sampling differs from common ray-sphere intersection due to differing observation directions.
We distribute sample points along each ray; we retain these samples throughout training and subsequent sphoxel-based samplings to prevent the optimization from failing to escape local minima.

\mparagraph{Coarse sphoxel sampling}
We select a set of coarse sphoxels $V_{\textup{coarse}}$ for sampling. Given the initial binoctree subdivision, we mark a sphoxel as $\in V_{\textup{coarse}}$ if it is a parent or grandparent of a leaf sphoxel.
Given the small baseline and large scene depth, this dilation step helps to resolve the sparse distribution of leaf sphoxels to better optimize plausible surface regions.
We compute ray entry and exit points for all marked sphoxels and uniformly sample in between~\cite{liu2020neural}.

One issue is the sky: if sampled, this background appearance will be optimized into the binoctree at an incorrect depth. Sky regions typically produce few points.  
Another is outlier points in the initial 3D point cloud. 
To handle both, we additionally prune sphoxels from $V_{\textup{coarse}}$ if the number of contained initial points is lower than a threshold, which varies per scene based on the point density from 3 to 20.

\begin{figure}[pt]
	\centering
	\includegraphics[width=1\linewidth]{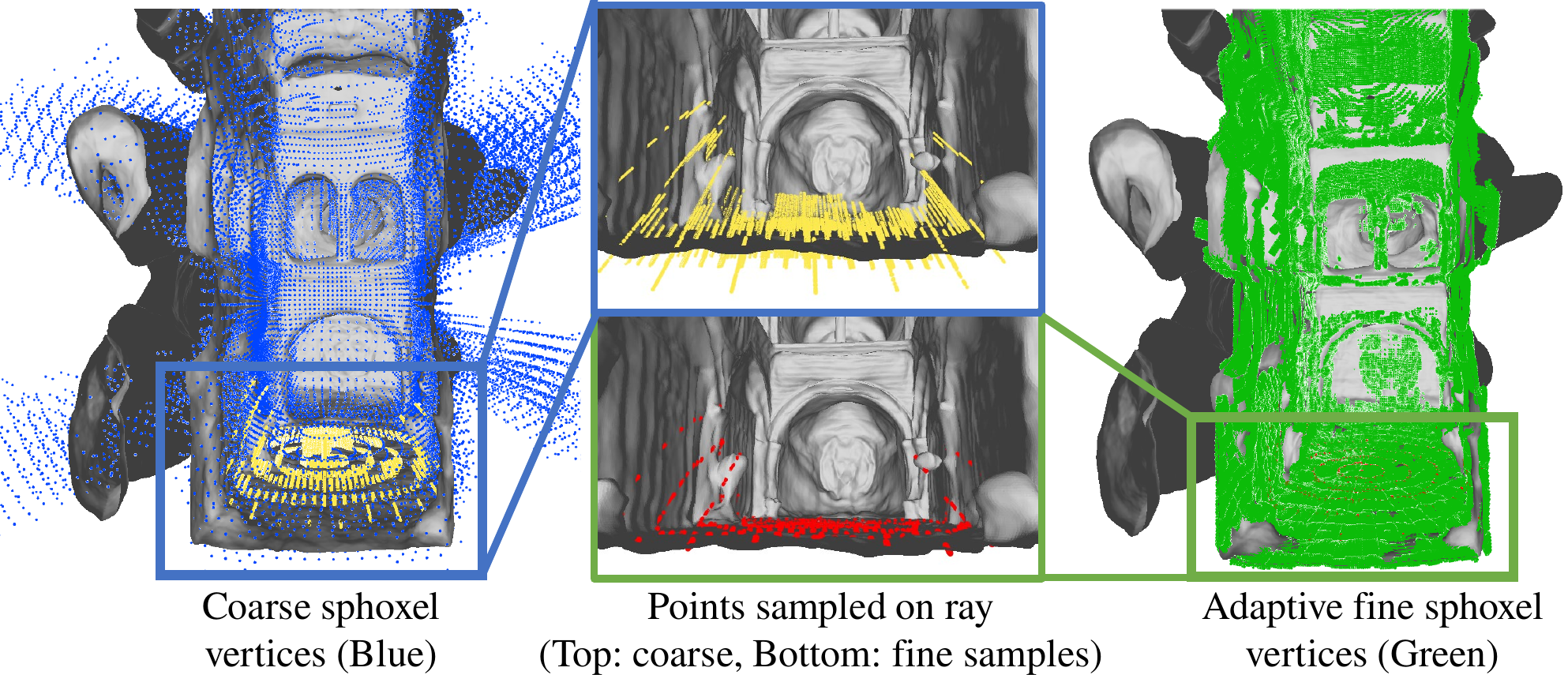}%
	\vspace{-2mm}
	\caption[Adaptive sampling]{
		Adaptive sampling sphoxel bounds.
			\emph{Left:} Coarse (blue) sphoxel vertices for sampling.
			\emph{Center:} A sample of points from coarse (top) and fine (bottom) sphoxel intersections.
			\emph{Right:} Fine (green) sphoxel vertices, which are denser near the surface.
	\label{fig:adaptive_sampling}}
	\vspace{-3mm}
\end{figure}

\mparagraph{Fine sphoxel sampling via adaptive subdivision}
Coarse sphoxel sampling provides a geometrical guide in the initial optimization iterations. 
However, sampling between all intersecting coarse sphoxels samples a larger region than the true surface location.
Therefore, to create $V_{\textup{fine}}$, we periodically reassess $V_{\textup{fine}}$ membership and then adaptively subdivide member sphoxels during optimization.
First, we initialize $V_{\textup{fine}}$ with $V_{\textup{coarse}}$. Then, as optimized SDF values may push a surface outside of $V_{\textup{fine}}$, we periodically fetch all leaf sphoxels in $V_{\textup{coarse}}$, infer the SDF $d$, and mark a sphoxel as being in $V_{\textup{fine}}$ if $\|d\|$ is smaller than each sphoxel's radius.
Then, we subdivided all marked sphoxels as $V_{\textup{fine}}$ such that surface optimization occurs at finer subdivisions.
Figure~\ref{fig:adaptive_sampling} shows sphoxel vertices in coarse and fine levels. We can see that adaptive subdivision detects the surface successfully and reduces the sampling range appropriately.

\subsection{Implementation Details}
\label{sec:details}

\mparagraph{Neural networks} We optimize three multi-linear perceptrons (MLPs): one each to decode the SDF and color features within the binoctree and one for the background. 
The SDF network has eight layers of 256 neurons, and the color network has four layers of 256 neurons. We use standard positional encoding with six frequency bands for the SDF network and ten for the background NeRF.

\mparagraph{Sample counts} We use 32 samples for each of the whole-space sphere, coarse voxel, and adaptive fine voxel stages. We used 32 samples for the background.
Within the sphoxels, we also conduct importance sampling along the ray based on the PDF along it, using another 32 samples.

\mparagraph{Subdivision schedule}
The binoctree is subdivided every 10\,k iterations until the leaf node size reaches a threshold angular size ranging 0.001--0.0001.

\mparagraph{Supervision} The rendered color is supervised with an L1 photometric loss, and the SDFs are supervised with an eikonal loss. 
We add a loss to manually mask out the tripod that supports the camera and to mask initial depth values near the epipoles where depth cannot be estimated. 
Therefore, the total loss is defined as:
\begin{align}\label{eq:loss}
\mathcal{L}_\text{total} = \mathcal{L}_\text{RGB} + \omega_\text{eik} \mathcal{L}_\text{eik}
+ \omega_\text{mask} \mathcal{L}_\text{mask}. 
\end{align}

\mparagraph{Iterations and computation} 
We optimize each scene on a computer equipped with a single NVIDIA A6000 GPU and an Intel Xeon Silver 4214R 2.40 GHz CPU with 256 GB RAM. It takes $\sim$20\,hours for 300\,k iterations. For the synthetic scenes, we evaluate the model after 100--200\,k iterations depending on their complexity: 200\,k for \emph{Sponza}, 200\,k for \emph{San Miguel}, and 150\,k for \emph{Lone-monk}.

\vspace{-0.1cm}
\section{Results}
\label{sec:results}
\vspace{-0.15cm}

\begin{figure*}[pt]
	\centering
	\includegraphics[trim=0mm 0mm 0mm 0mm, clip, width=\linewidth]
	{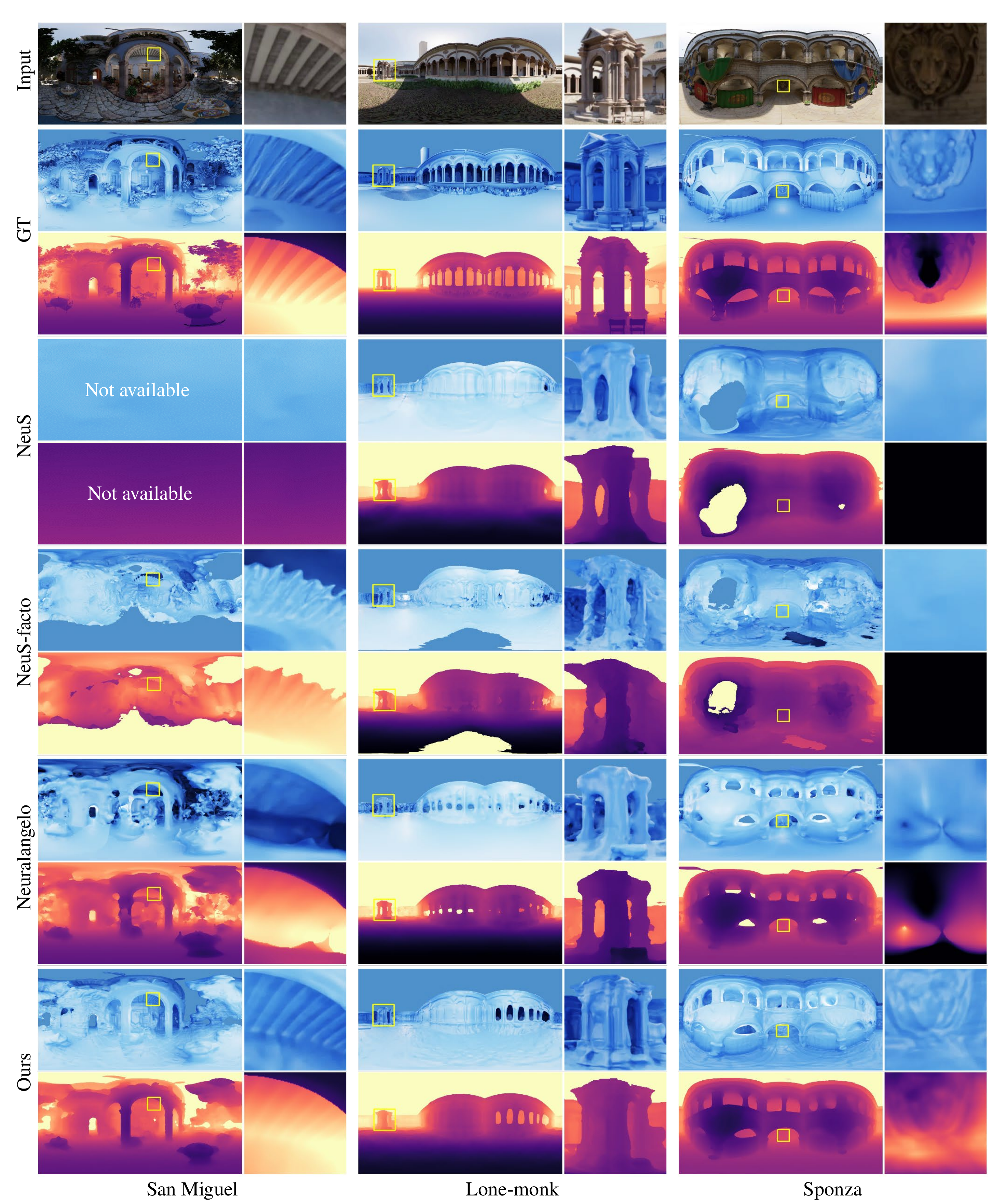}%
	\vspace{-3mm}%
	\caption[Sampling strategies]{\label{fig:synthetic_result}
		We compare our method with traditional and neural methods using ground-truth geometry. We present qualitative results of NeuS~\cite{wang2021neus}, NeuS-facto~\cite{Yu2022SDFStudio} and Neuralangelo~\cite{li2023neuralangelo} here; our method produce higher-quality 3D geometry. Please refer to Table~\ref{tab:eval_synthetic} for complementary quantitative evaluation and to supplemental material for further qualitative comparisons of omitted traditional methods, including COLMAP~\cite{schonberger2016structure}, and EgocentricRecon~\cite{jang2022egocentric}.}
\end{figure*}

\mparagraph{Evaluation Method}
We compare to five methods for reconstructing omnidirectional images.
First, we use the classical structure-from-motion method COLMAP~\cite{schonberger2016structure} after converting omnidirectional images to a six-face perspective cube map. 
Second, we compare to EgocentricRecon~\cite{jang2022egocentric}, an omnidirectional reconstruction method designed for the same kind of input as ours, and also using a binoctree.
Third, fourth, and fifth, we compare our reconstruction results with three neural surface reconstruction methods: original NeuS~\cite{wang2021neus},  NeuS-facto~\cite{wang2021neus,Yu2022SDFStudio}, and Neuralangelo~\cite{li2023neuralangelo}.
NeuS-facto is a neural SDF reconstruction method that adapts NeuS by using the proposal network from mip-NeRF360 for sampling points along the ray. 
This comparison is suitable for unbounded scenes, whereas the original NeuS does not support these.
Neuralangelo trains coarse to fine hash table grids with numerical gradients for higer-order derivatives.%
We use each method's implementation from SDF studio~\cite{Yu2022SDFStudio}.
Approximate runtime took 10 hours for COLMAP, 30 minutes for EgocentricRecon, 5.5 hours for NeuS, 25 minutes for NeuS-Facto, and 31 hours for Neuralangelo to process the \emph{Sponza} scene (Fig.~\ref{fig:synthetic_result}) with 200 images.
Our method took 12.5 hours.

\mparagraph{Dataset} We use synthetic scenes for ground truth quantitative evaluation, using 3D scenes from the Blender Online Community and the McGuire Computer Graphics Archive. We render equirectangular RGB-D video with Blender at 2048 $\times$ 1024 pixels. The camera has a circular trajectory within a central region inside the scenes, spanning 200 frames.
These three scenes---\emph{Sponza}, \emph{Lone-monk}, and \emph{San Miguel}---have fine detail, especially \emph{San Miguel}, which has many detailed trees that are challenging to reconstruct. We also show real-world reconstruction from the Omniphotos dataset captured with a swinging selfie stick~\cite{bertel2020omniphotos}.

\mparagraph{Space subdivision efficiency}
Our binoctree subdivision is an efficient method to reduce the number of voxels and tree depths when compared to the naive Cartesian subdivision. The number of uniform Cartesian voxels required to fill a unit sphere with the same size as the smallest sphoxel in our binoctree is great (Table~\ref{tab:eval_memory}): our spherical subdivision requires about 10\,k $\times$ fewer voxels.

\begin{table}[tp]
	\centering
	\caption{\label{tab:eval_memory}	
The number of voxels in each scene for a Cartesian subdivision grid is more than in our adaptive sphoxel grid. Our method is significantly more efficient while still achieving high-frequency details in reconstructed results.}
	\vspace{-3mm}
	\resizebox{1.0\linewidth}{!}{%
		\begin{tabular}{ll rr}
			\toprule
			Scene & 	Method	& Number of voxels & Minimum sphoxel size \\
			\midrule
			\multirow{2}{*}{Sponza} & Dense regular grid & 33,335,054,331  & 	  \\
									& Ours			  		& 4,346,041 & 1.25e-10 \\
			\multirow{2}{*}{Lone-monk} & Dense regular grid & 2,234,638,740 &  \\
									& Ours			  & 231,237 & 1.87e-9 \\
			\multirow{2}{*}{San Miguel} & Dense grid & 1,953,273,076 &  \\
									& Ours			  & 951,703 & 2.14e-9 \\
			\bottomrule
		\end{tabular}
	}
	\vspace{-3mm}
\end{table}

\mparagraph{Surface reconstruction accuracy}
\begin{table}[pt]
	\centering
	\caption{\label{tab:eval_synthetic} Quantitative evaluation of surface mesh accuracy for both classical and neural methods. We measure MAE and RMSE by rendering the inverse depth of each mesh and comparing to the ground truth mesh. Methods in bold achieve the best accuracy.}
		\vspace{-3mm}
	\resizebox{\linewidth}{!}{
		\begin{tabular}{lcccccccc}
			\toprule
			& \multicolumn{2}{c}{Sponza} & \multicolumn{2}{c}{Lone-monk} &    \multicolumn{2}{c}{San Miguel} & \multicolumn{2}{c}{Average} \\
			\cline{2-9}
			&RMSE   &MAE    &RMSE   &MAE    &RMSE   &MAE    &RMSE   &MAE\\
			\midrule
			COLMAP  				 &0.88   &0.28   &\bf{0.40}   &\bf{0.15}   &1.24   &0.41   &0.84   &\bf{0.28}\\
			EgocentricRecon		 &\bf{0.75}   &\bf{0.27}   &1.05   &0.43   &\bf{0.82}   &\bf{0.29}   &0.88   &0.33\\
			\midrule
			NeuS 						 &2.88   &1.43   &1.39   &1.24   & --    & --    &2.13   &1.33\\
			Neus-facto 				  &2.44   &1.12   &2.25   &1.68   &4.02   &2.84   &2.91   &1.88\\
			Neuralangleo 			&0.99   &0.44   &1.91   &1.71   &0.58   &0.29   &1.16   &0.81\\
			\midrule
			Ours					&\bf{0.82}   &\bf{0.35}   &\bf{0.69}   &\bf{0.36}   &\bf{0.56}   &\bf{0.27}   &\bf{0.69}   &\bf{0.32}\\
			\bottomrule
		\end{tabular}
	}
		\vspace{-3mm}
\end{table}
Since only a portion of the complete ground truth mesh is reconstructed within the omnidirectional video sweep setting, a direct mesh accuracy metric such as Hausdorff distance is inappropriate for evaluation.
Instead, to evaluate the accuracy of surface reconstruction, we render the depth of the reconstructed mesh from the center of the camera trajectory for all methods. Then, we compare each to the rendered depth of the ground truth mesh from the same position.

We measure the depth MSE and RMSE of valid (e.g., non-sky) ground truth mesh surfaces in Table~\ref{tab:eval_synthetic}. 
Our approach can optimize an SDF with comparable results to classical reconstruction methods while greatly exceeding the reconstruction quality of existing neural SDF methods.
NeuS-facto shows improved accuracy in the Sponza scene and reconstructs the surface of San Miguel, which NeuS completely fails to reconstruct.
As such, we leave out the San Miguel reconstruction result of NeuS since the reconstruction fails to create a mesh and so we cannot align and measure its accuracy. 
Neuralangelo shows the best reconstruction results among previous neural methods.
However, our method achieves a higher accuracy still.	
We provide the qualitative comparison with neural methods in Figure~\ref{fig:synthetic_result}, and with classical methods in the supplemental material.
This validates that our method can successfully reconstruct detailed surfaces of large and complex outdoor scenes that current neural methods could not---simple methods based only on MLPs cannot easily scale to large scenes.

\begin{figure*}[pt]
\centering
\vspace{-4mm}%
\includegraphics[trim=0mm 0mm 0mm 0mm, clip, width=0.9\linewidth]{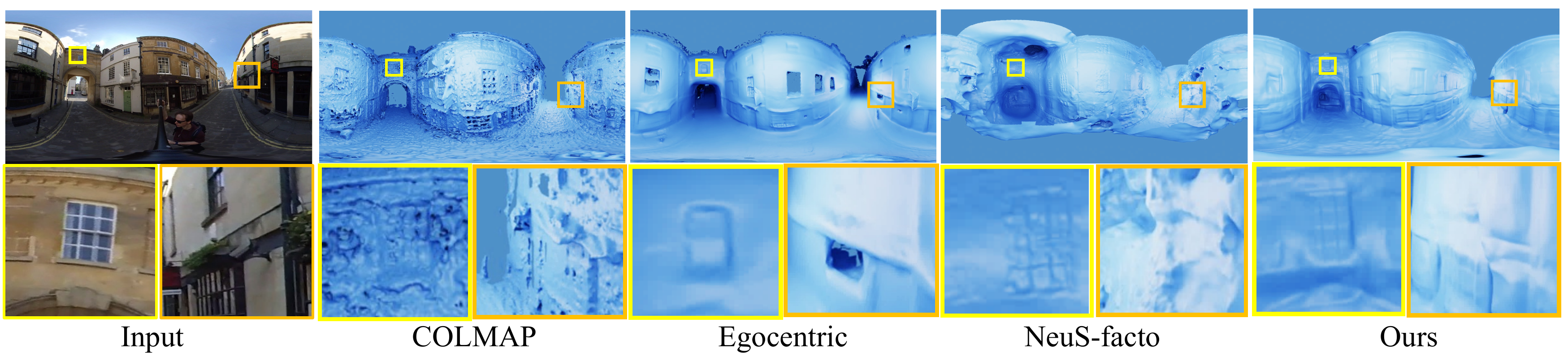}%
\vspace{-3mm}%
\caption[Sampling strategies]{\label{fig:real_result}
	Qualitative comparisons of our reconstruction on a real-world scene from the Omniphotos dataset. Please refer to the supplemental video for more results.}	
\end{figure*}

Figure~\ref{fig:real_result} shows our qualitative reconstruction accuracy on a real scene. Our method can reconstruct both the scene details and smooth surfaces within the geometry. COLMAP reconstructs details without filling in holes; EgocentricRecon is overly smooth, and NeuS-facto is often in error.

\begin{figure*}[thp]
\vspace{-4mm}%
\hspace{3mm}%
\includegraphics[trim=0mm 0mm 0mm 0mm, clip, width=0.93\linewidth]{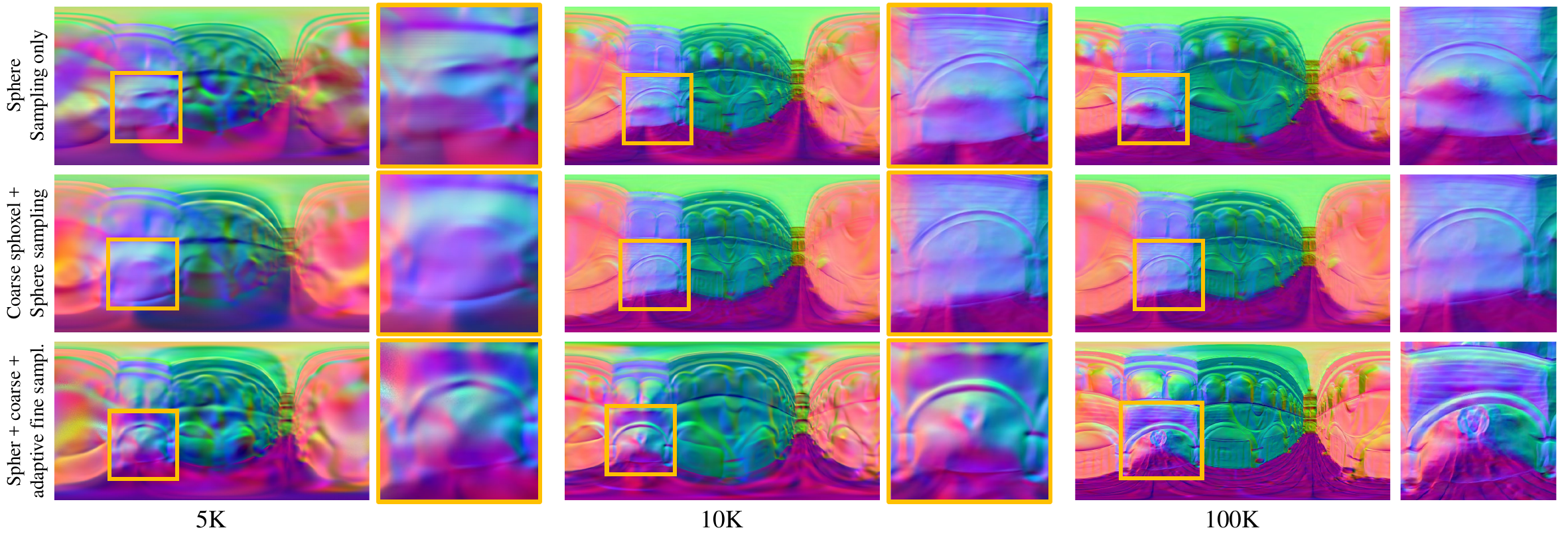}%
\vspace{-3mm}%
\caption[Iteration progress]{
	As an ablation, we show the progress of geometry optimization with the omnidirectional normal map of the \emph{sponza} scene. Normal maps are rendered per training iterations of 5K, 10K, and 100K for each sampling strategy. The geometry in the hollow region was recovered only when both coarse and fine sampling was adopted.
	\label{fig:sampling_ablation}}
	\vspace{-4mm}%
\end{figure*}

\mparagraph{Comparison with Depth Supervision}
We compare our approach, which uses a binoctree pre-built from depth input as a sampling guide, to NeuS-facto-D, a method that uses depth supervision directly to guide surface reconstruction. 
Our findings suggest that using the binoctree as a guide is superior to direct depth supervision.
To incorporate depth supervision, NeuS-facto-D adds a depth loss component to the loss with a weight of 0.1. 
The depth loss is calculated using L1 loss between the rendered depth and input depth.
Depth supervision can improve surface reconstruction on average but our approach yields more significant improvements (\Cref{tab:suppl_depth_supervision}). 
One reason why is because depth supervision can often be inaccurate. 
Our approach conservatively sets bounds upon the depth and then relies upon other losses to optimize the surface location.

\begin{table}[tp]
	\centering
	\caption{\label{tab:suppl_depth_supervision} Quantitative comparison of binoctree-based geometry guidance and depth supervision for geometry reconstruction quality evaluation using RMSE depth accuracy.}	
	\vspace{-3mm}
	\resizebox{1.0\linewidth}{!}{%
		\begin{tabular}{lcccc}
			\toprule
			Method &    Sponza & Lone-monk & San Miguel  & Average \\
			\midrule
			NeuS-facto~\cite{Yu2022SDFStudio}               &2.44   &2.25   &4.03   &2.91   \\
			NeuS-facto-D~\cite{Yu2022SDFStudio}             &1.46   &4.75   &1.80    &2.67 \\
			\midrule
			Ours                                             &\textbf{0.82}   &\textbf{0.69}   &\textbf{0.56}   &\textbf{0.69} \\
			\bottomrule
		\end{tabular}
	}
	\vspace{-4mm}
\end{table}

\mparagraph{Ablation}
To demonstrate the effectiveness of our sampling method, we display the training progress using normal maps for three different approaches: 
sphere sampling only, sphere and coarse sphoxel sampling, and hybrid (sphere, coarse, and fine sphoxel) sampling 
at 5\,k, 10\,k, and 100\,k training iterations. 
For fairness, we keep the total number of samples per ray constant at 128.
Our baseline setting (sphere sampling only) follows the default sampling strategies from NeuS except that it uses outward-looking sphere sampling instead(Fig~\ref{fig:sampling_types} (b)). Then, we sample more between coarse sphoxels instead of sphere bounds. Lastly, we sample fine sphoxels instead of importance sampling.
Table~\ref{tab:ablation} shows the number of samples used for this ablation. 

\begin{figure}[pt]
	\centering
	\includegraphics[trim=0mm 70mm 0mm 0mm, clip, width=\linewidth]{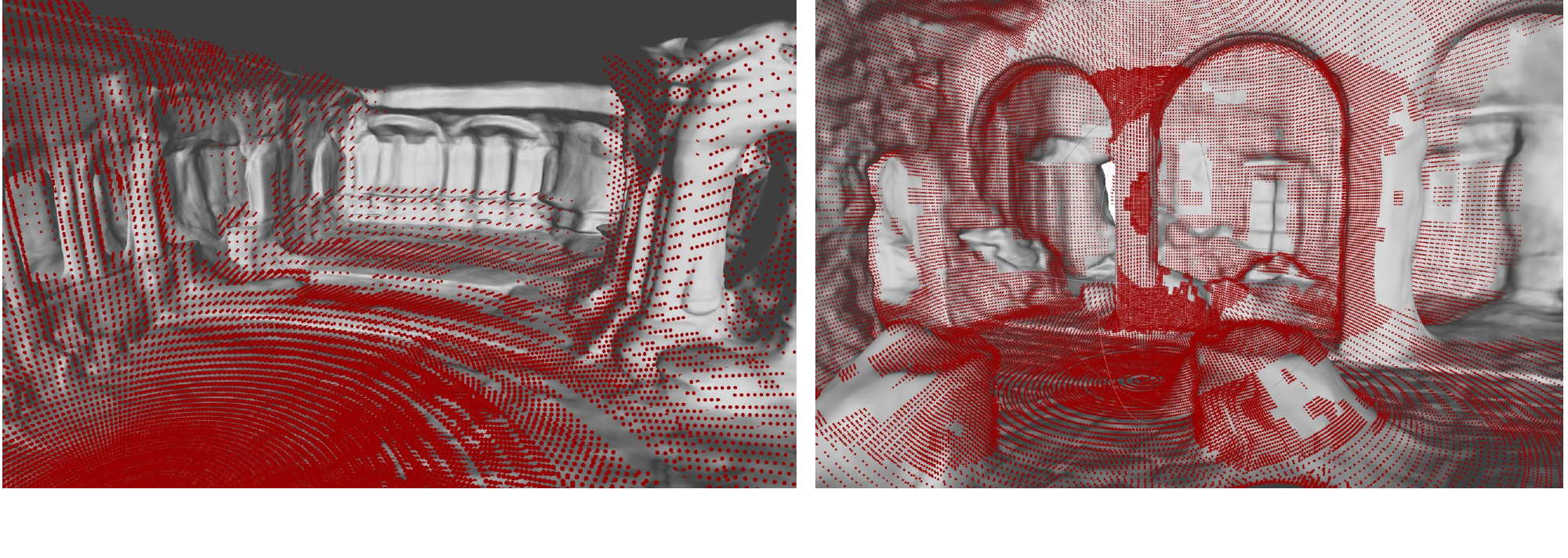}%
	\vspace{-5mm}%
	\caption[Synthetic dataset]{\label{fig:surface_sphoxel_convergence}
		Adaptive fine voxel bounds (red point samples) have correctly converged to the surface location.}
	\vspace{-4mm}%
\end{figure}

Figure~\ref{fig:sampling_ablation} (bottom right) shows that scene details in concave regions are estimated with accuracy only if ray sampling is strongly guided by both coarse and fine sphoxels. 
Figure~\ref{fig:surface_sphoxel_convergence} qualitatively demonstrates that our adaptive fine voxel correctly converges to the surface location.

\begin{table}[tp]
	\centering
	\caption{\label{tab:ablation} Number of samplings for sampling strategies ablations}
	\vspace{-3mm}
	\resizebox{1.0\linewidth}{!}{%
		\begin{tabular}{lcccc}
			\toprule
			Sampling techniques & 	$N_\text{sphere}$.	& $N_\text{coarse}$ & $N_\text{fine}$  & $N_\text{importance}$ / steps \\
			\midrule
			Sphere sampling							& 64	& 0	   & 0		& 64 / 4 \\
			Sphere + coarse voxel				  & 32	  & 32  & 0		 & 64 / 4 \\
			Sphere + coarse + fine voxel		& 32	& 32   &   32	  & 32 / 2 \\
			\bottomrule
		\end{tabular}
		}
\end{table}

\vspace{-0.1cm}
\section{Discussion and Conclusions}
\label{sec:discussion}
\vspace{-0.15cm}

We have presented an adaptive subdivision of spherical space via a binoctree with novel sampling techniques. This subdivision and sampling are useful to optimize a neural implicit representation of geometry from an omnidirectional video. When such a representation is optimized for large unbounded scenes, the continuous property of the representations often causes smoothed surfaces that lack detail. To overcome this issue, we use an adaptive subdivision strategy that densifies samples near the surface. This strategy subdivides the sphoxel towards the surface and moves samples similarly so that density is placed where it is needed. 

To address the challenges posed by using a neural function to estimate abrupt changes in disparity and sparse observations from omnidirectional input, we propose constructing coarse space bounds from depth input to provide a geometrical guide during optimization. The spherical binoctree efficiently subdivides the omnidirectional space, taking into account the rendering resolution and memory space. We subdivide the space based on solid angle measurement and distance from the camera, using a spatially variant tree depth, using orders of magnitude fewer voxels than a Cartesian grid with an equivalent minimum voxel size. This leads to improved surface detail without requiring significant memory. Our model performance is comparable to traditional surface reconstruction methods; concurrent neural methods using MLPs alone cannot achieve this quality for our inputs, according to our comparisons. Overall, quantitative, qualitative, and ablation experiments confirm the effectiveness of our approach as a promising method for small-baseline omnidirectional data.

\mparagraph{Limitations}
Our approach uses an initial depth map to initialize the coarse voxel before subdividing them into finer levels. Should the estimation of the coarse voxels be wrong, it can lead to inaccurate guidance for the coarse geometry. This also may cause inaccuracies when samples are estimated at scene regions that should belong to the background but have nearer initial depths. We attempt to lessen these effects by dilating and pruning the voxel before optimization, but it may be necessary to perform postprocessing to ensure better final meshes without spurious artifacts.

Further, some scene geometry details still remain challenging to reconstruct, e.g., concave region reconstruction may be sensitive to particular minima in the optimization landscape. Further, effects that should be explained by fine-scale texture variation may be baked into the geometry.

\appendix
\vspace{-0.1cm}
\section*{Acknowledgements}
\vspace{-0.15cm}
\noindent Min H.~Kim acknowledges the MSIT/IITP of Korea (RS-2022-00155620, 2022-0-00058, and 2017-0-00072), LIG, and Samsung Electronics. 
James Tompkin acknowledges US NSF CAREER 2144956.

\clearpage
{
    \small
    \bibliographystyle{ieeenat_fullname}
    \bibliography{spherical-3d-recon}
}

	\clearpage
	\setcounter{page}{1}
	\maketitlesupplementary
	
	\IGNORE{

\begin{mdframed}
Dear reviewers,\\

\noindent 
We accidentally reported RMSE as MSE in Table 1 of our main paper. It was an unintentional mistake and we apologize for any confusion. In supplemental Table~\ref{tab:suppl_depth_accuracy}, we separately report both MSE and RMSE, along with MAE.

We would also like to draw your attention to three additional points:
\begin{itemize}
	\item Along with NeuS-facto, we report metrics for the original NeuS and newer Neuralangenlo methods.
	\item We initially reported a higher NeuS-facto MSE due to an incorrect mesh scale, used during metric computation only. This has been corrected. While NeuS-facto is now more accurate by MSE, it is still less accurate than existing classical COLMAP, SOTA EgocentricRecon, and our approach. 
	\item The qualitative results remain the same.\\
\end{itemize}
\noindent Thank you for your understanding and consideration.\\[3mm]
\noindent Sincerely,\\
\noindent The authors
\end{mdframed}
}

\section{Additional Qualitative Comparisons}
\begin{figure*}[t]
	\centering
	\includegraphics[trim=0mm 0mm 0mm 0mm, clip, width=\linewidth]{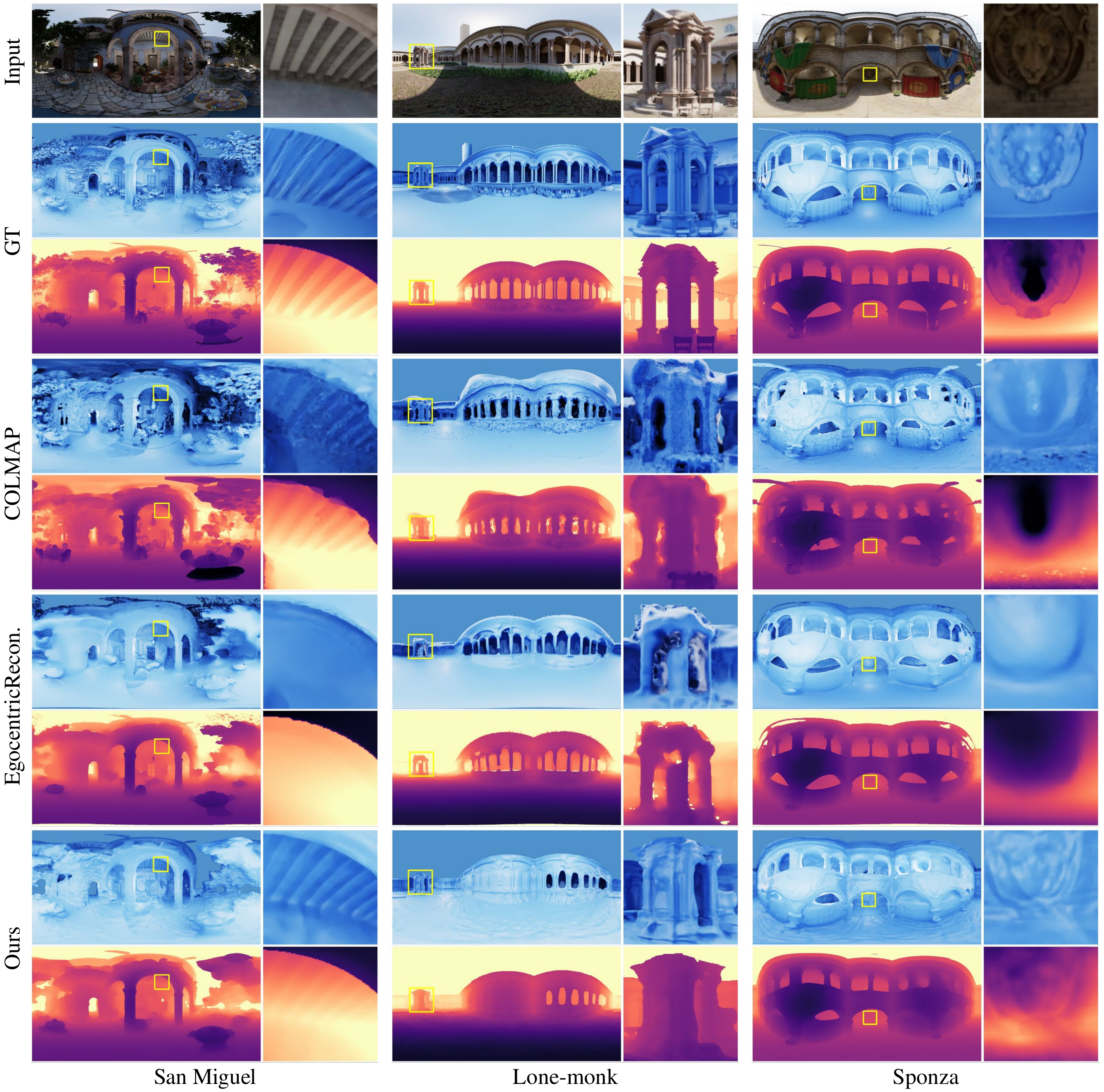}%
	\caption[Sampling strategies]{\label{fig:supple_synthetic}
		We compare our method with traditional and neural methods using ground-truth geometry, including COLMAP~\cite{schonberger2016structure}, and EgocentricRecon~\cite{jang2022egocentric}.
		We also compare 3D reconstructed geometry rendering and depth maps and observe that our method produced higher-quality 3D geometry than EgocentricRecon and smoother reconstruction than COLMAP.}
\end{figure*}

We present qualitative comparisons of our reconstruction results on synthetic scenes with two traditional surface reconstruction methods: COLMAP~\cite{schonberger2016structure} and EgocentricRecon~\cite{jang2022egocentric} in Figure~\ref{fig:supple_synthetic}.
The quantitative and qualitative results of traditional methods are comparable to ours.
However, Figure~\ref{fig:supple_synthetic} shows that in some areas our method can recover details and smooth surfaces that traditional methods do not.

\begin{figure*}[t]
	\centering
	\vspace{-3mm}%
	\includegraphics[trim=0mm 0mm 0mm 0mm, clip, width=0.9\linewidth]{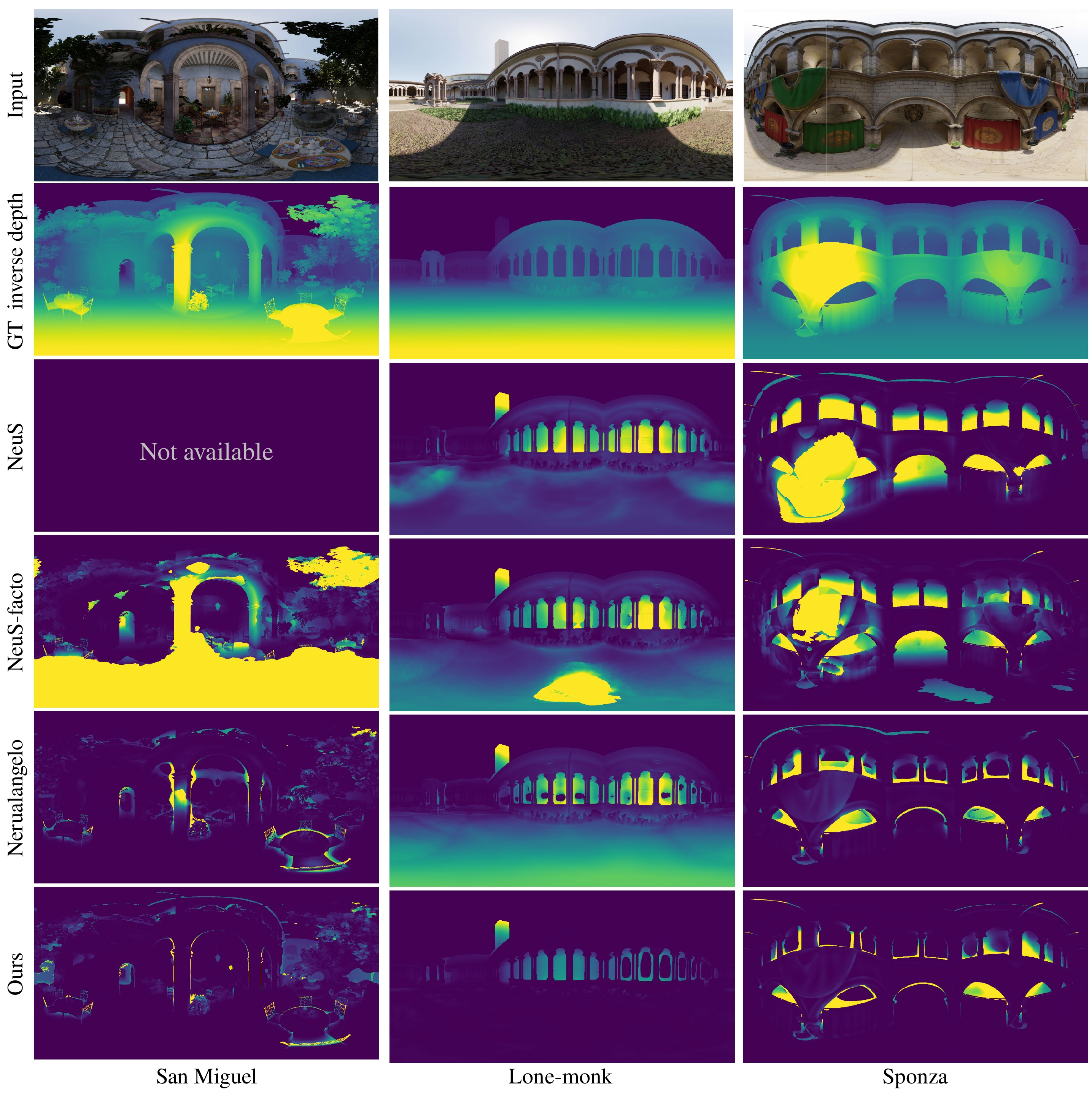}%
	\vspace{-3mm}%
	\caption[Sampling strategies]{\label{fig:supple_errormap}
Comparison of error maps between our methods and other neural methods. While all approaches have some errors, our approach generally has lower errors. NeuS fails to execute on the San Miguel scene.}
	\vspace{1mm}
	
\end{figure*}

We also include omitted qualitative comparisons on the real scenes from our main paper in Figures~\ref{fig:supple_real_comparison}. 
We also provide error maps for all neural methods to compare error distributions better. 
Figure~\ref{fig:supple_errormap} shows that our method can better estimate regions with abrupt disparity changes, for example, the concave area behind the pillars.

\section{Additional Real Scene Results}
\begin{figure*}[t]
	\centering
	\includegraphics[trim=0mm 0mm 0mm 0mm, clip, width=0.9\linewidth]{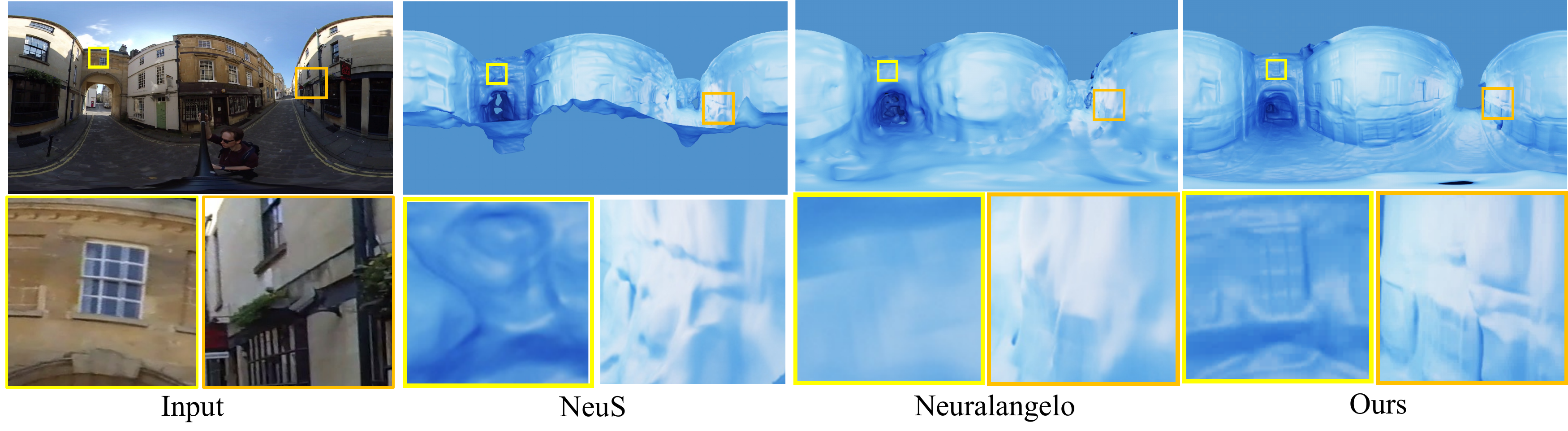}%
	\caption[Sampling strategies]{\label{fig:supple_real_comparison}
		Additional real-scene qualitative comparisons of our method with 	NeuS~\cite{wang2021neus} and Neuralangelo~\cite{li2023neuralangelo}.}
	
	\centering
	\includegraphics[trim=0mm 0mm 0mm 0mm, clip, width=\linewidth]{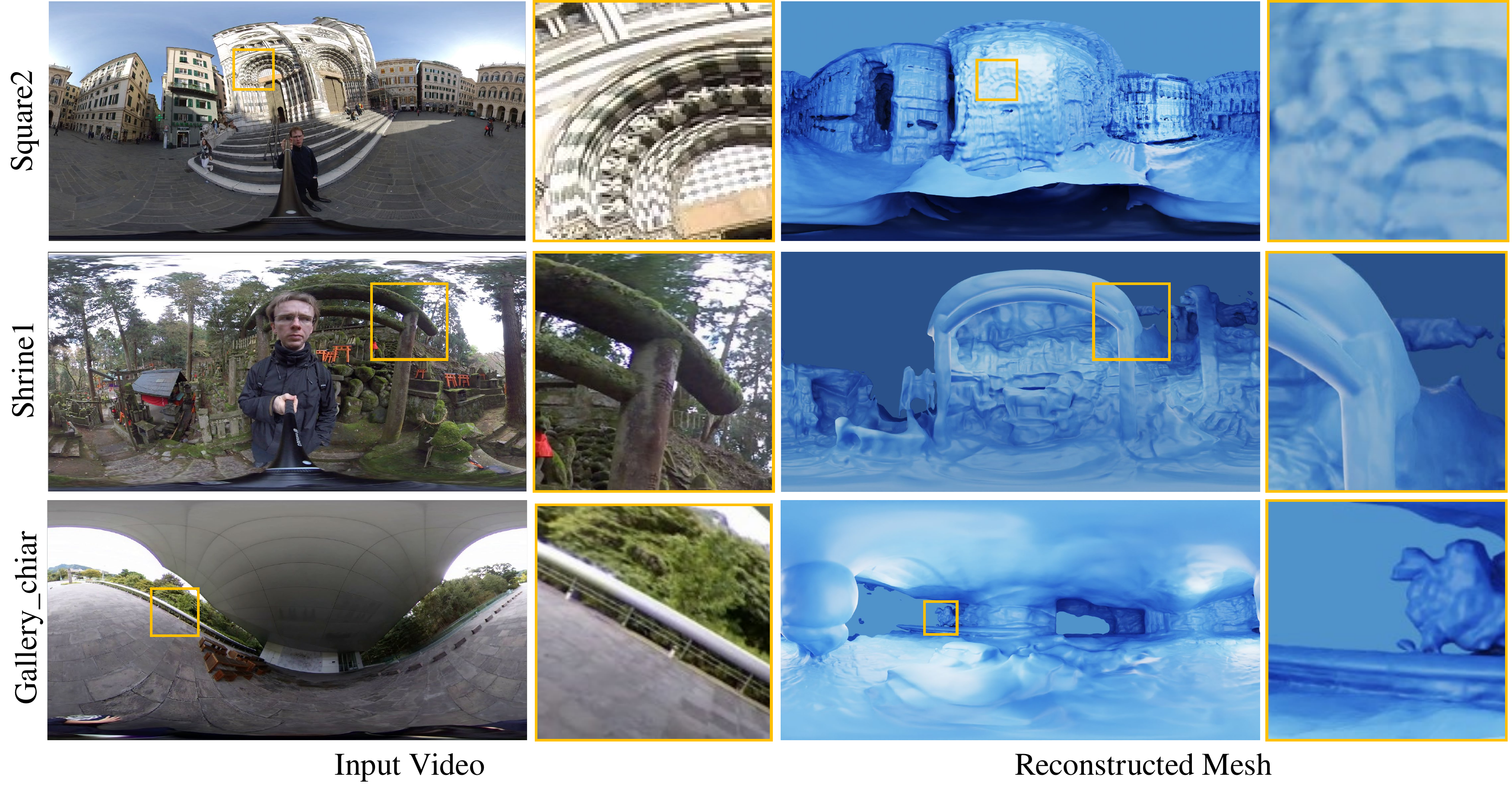}%
	\caption[Sampling strategies]{\label{fig:supple_real_additional}
		Additional reconstruction results from real-scene circular sweep baseline videos for our method.}
\end{figure*}

We present additional results of real scenes, including `gallery chair' from Richo360~\cite{choi2023balanced}, and `Shrine 1' and `Square 2' from the Omniphotos~\cite{bertel2020omniphotos} dataset (\Cref{fig:supple_real_additional}).

\begingroup
\raggedright
\small
\def\addvspace#1{\vspace{4pt}}
{
    \small
    \bibliographystyle{ieeenat_fullname}
    \bibliography{spherical-3d-recon}
}
\endgroup

\end{document}